%% file: main.tex
\documentclass[conference]{IEEEtran}
\IEEEoverridecommandlockouts
\usepackage{times}
\usepackage[numbers]{natbib}
\usepackage{multicol}
\usepackage[bookmarks=true]{hyperref}
\usepackage{color}
\usepackage{textcomp}
\usepackage{amsmath,amsfonts,amssymb}
\usepackage{graphicx}
\usepackage{makecell}
\usepackage{tabularx}
\usepackage{graphicx}
\usepackage{pgfplots}
\usepackage{tikz}
\usepackage{array}
\usepackage{subcaption}
\usepackage{cleveref}
\usepackage[linesnumbered,ruled,norelsize]{algorithm2e}
\newcolumntype{?}{!{\vrule width 1pt}}

\definecolor{bblue}{HTML}{4F81BD}
\definecolor{rred}{HTML}{C0504D}

\hypersetup{
  colorlinks   = true, 
  urlcolor     = blue, 
  linkcolor    = black, 
  citecolor   = black 
}

\tikzset{level 1/.style={level distance=0.7cm, sibling distance=1.0cm}}
\tikzset{level 2/.style={level distance=0.7cm, sibling distance=4.0cm}}
\tikzset{bag/.style={draw, rectangle, solid, text width=7em, text centered, yshift=-0.2cm}}

\newcommand{\ouralgo}{POMDSoar\xspace}
\newcommand{\suav}[1]{sUAV#1\xspace}
\newcommand{\extradata}[1]{Appendix}

\makeatletter
\newcommand{\removelatexerror}{\let\@latex@error\@gobble}
\makeatother

\newcommand{\comment}[1]{}

\input{./MathDefs.tex}

\title{Autonomous Thermalling as a Partially Observable Markov Decision Process\\ (Extended Version)}

\author{\authorblockN{Iain Guilliard\thanks{The author did most of the work for this paper while at Microsoft Research.}}
\authorblockA{Australian National University\\ Canberra, Australia\\iainguilliard@gmail.com}
\and
\authorblockN{Richard Rogahn}
\authorblockA{Microsoft Research\\
Redmond, WA-98052\\
rrogahn@microsoft.com}
\and
\authorblockN{Jim Piavis}
\authorblockA{Microsoft Research\\
Redmond, WA-98052\\
v-jimpi@microsoft.com}
\and
\authorblockN{Andrey Kolobov}
\authorblockA{Microsoft Research\\
Redmond, WA-98052\\
akolobov@microsoft.com}
}

\begin{document}
\newcommand{\backupsmall}{{\vspace*{-0.2cm}}}

\maketitle
\input{abstract}
\input{introduction}
\input{background}
\input{problem_formalization}

\input{algorithm}

\input{related_work}
\input{experiments}

\input{conclusions}

\input{appendix}

\bibliographystyle{plainnat}
\bibliography{bibliography}
\end{document}

%% file: MathDefs.tex
\usepackage{xspace}

\newcommand{\thmlstate}{\ensuremath{s^{th}}\xspace}
\newcommand{\uavstate}{\ensuremath{s^{u}}\xspace}
\newcommand{\thmlpos}{\ensuremath{\vec{p}^{th}}\xspace}
\newcommand{\uavpos}{\ensuremath{\vec{p}^{u}}\xspace}
\newcommand{\yaw}{\ensuremath{\psi}\xspace}
\newcommand{\roll}{\ensuremath{\phi}\xspace}
\newcommand{\rollrate}{\ensuremath{\dot{\phi}}\xspace}

\newcommand{\alt}{\ensuremath{h}\xspace}
\newcommand{\airspeed}{\ensuremath{v}\xspace}
\newcommand{\thmlz}{\ensuremath{W_0}\xspace}
\newcommand{\thmlr}{\ensuremath{R_0}\xspace}
\newcommand{\Iroll}{\ensuremath{I_x}\xspace}
\newcommand{\gravity}{\ensuremath{g}\xspace}
\newcommand{\aileronout}{\ensuremath{\delta_a}\xspace}
\newcommand{\Clp}{\ensuremath{\mathbf{C}_{lp}}\xspace}
\newcommand{\LPdamp}{\ensuremath{L_p}\xspace}
\newcommand{\Kdamp}{\ensuremath{\mathbf{K}_d}\xspace}
\newcommand{\Kaileron}{\ensuremath{\mathbf{K}_a}\xspace}
\newcommand{\rollaction}{\ensuremath{\phi_a}\xspace}

%% file: abstract.tex
\begin{abstract}
Small uninhabited aerial vehicles (\suav{s}) commonly rely on active propulsion to stay airborne, which limits flight time and range. To address this, autonomous soaring seeks to utilize free atmospheric energy in the form of updrafts (thermals). However, their irregular nature at low altitudes makes them hard to exploit for existing methods. We model autonomous thermalling as a POMDP and present a receding-horizon controller based on it. We implement it as part of ArduPlane, a popular open-source autopilot, and compare it to an existing alternative in a series of live flight tests involving two \suav{s} thermalling simultaneously, with our POMDP-based controller showing a significant advantage.
\end{abstract}

%% file: introduction.tex
\section{Introduction}

Small uninhabited aerial vehicles (\suav{s}) commonly rely on active propulsion stay in the air. They use motors either directly to generate lift, as in copter-type \suav{s}, or to propel the aircraft forward and thereby help produce lift with airflow over the drone's wings. Unfortunately, motors' power demand significantly limits \suav{s'} time in the air and range. 

\begin{figure}[t]
\centering
\includegraphics[width=0.45\textwidth]{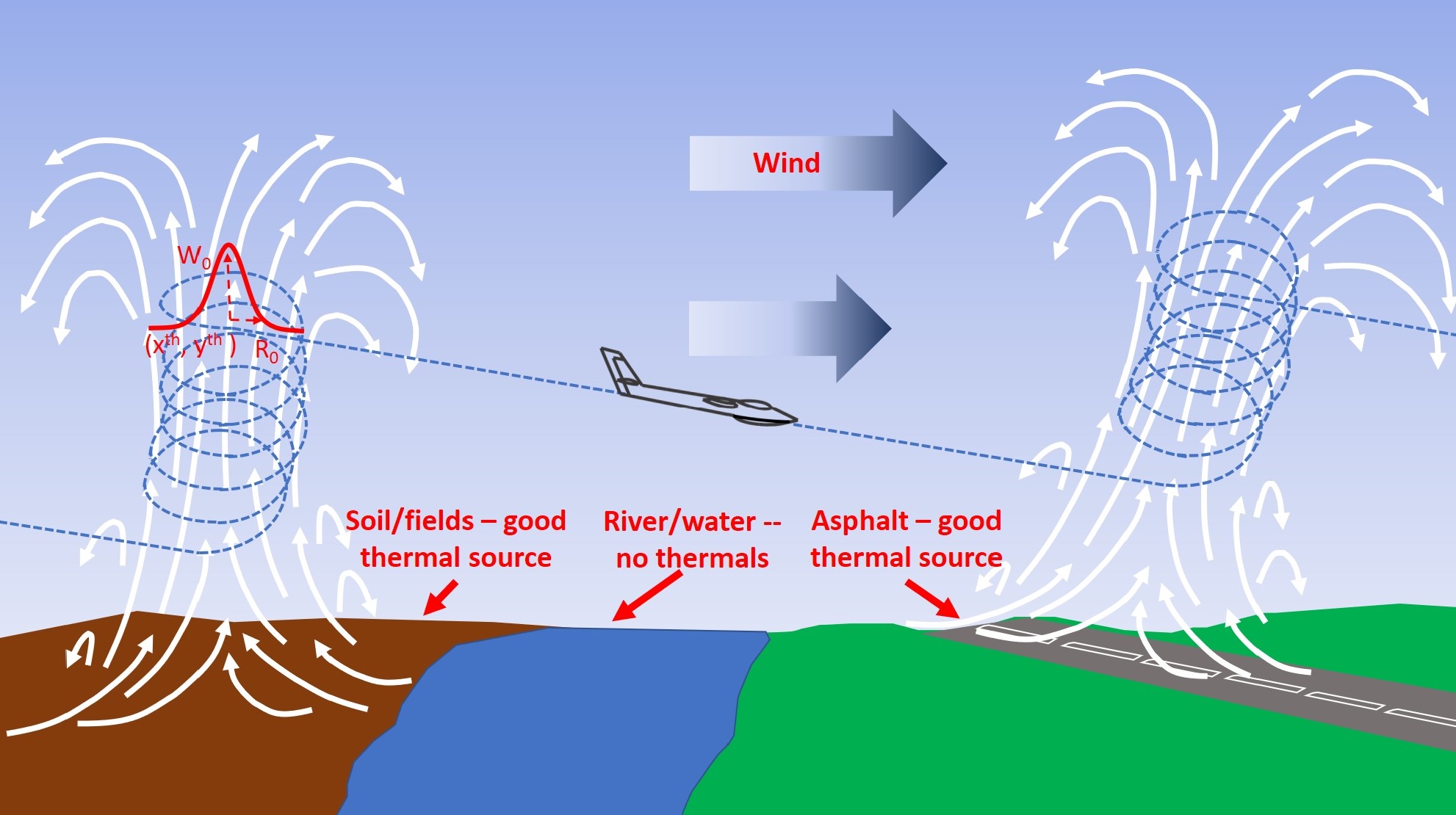}
\caption{Thermals and their bell-shaped lift model (in red).}
\label{fig:thermals}
\vspace{-0.7cm}
\end{figure}

In the meantime, the atmosphere has abundant energy sources that go unused by most aircraft. Non-uniform heating and cooling of the Earth's surface creates \emph{thermals} --- areas of rising air that vary from several meters to several hundreds of meters in diameter (see Figure \ref{fig:thermals}). Near the center of a thermal air travels upwards at several meters per second. Indeed, thermals are used by human sailplane pilots and many bird species to gain hundreds of meters of altitude \cite{akos-bb10}. A simulation-based theoretical study estimated that under the exceptionally favorable thermalling conditions of Nevada, USA and in the absence of altitude restrictions, an aircraft's 2-hour endurance could potentially be extended to 14 hours by exploiting these atmospheric updrafts \cite{allen-aiaa05,allen-aiaa06}.

Researchers have proposed several approaches to enable autonomous thermalling for fixed-wing \suav{s} \cite{wharington-1998,wharington-icas98,allen-2007,edwards-aiaa2008,hazard-aiaa10,reddy-pnas16,lecarpentier-jfpda17}. Generally, they rely on various parameterized thermal models characterizing vertical air velocity distribution within a thermal. In order to successfully gain altitude in a thermal, a \suav{'s} autopilot has to discover it, determine its parameters such as shape and lift distribution inside it, construct a trajectory that would exploit this lift, and exit at the right time. In this paper, we focus on autonomously identifying thermal parameters and using them to gain altitude. These processes are interdependent: thermal identification influences the choice of trajectory, which, in turn, determines what information will be collected about the thermal; both of these affect the decision to exit the thermal or stay in it.

Reinforcement learning (RL) \cite{sutton-98}, a family of techniques for resolving such exploration-exploitation tradeoffs,  has been considered in the context of autonomous thermalling, but only in simulation studies \cite{wharington-1998,wharington-icas98,reddy-pnas16,lecarpentier-jfpda17}. Its main practical drawback for this scenario is the \emph{episodic} nature of classic RL algorithms. RL agents learn by executing sequences of actions (episodes) that are occasionally ``reset'', teleporting the agent to its initial state. If  the agent has access to an accurate resettable simulator of the environment, this is not an issue, and there have been attempts to build sufficiently detailed thermal models \cite{reddy-pnas16} for learning thermalling policies offline. However, to our knowledge, policies learned in this way have never tested on real \suav{s}. Lacking a highly detailed simulator, in order to learn a policy for a specific thermal, a \suav{} would need to make many attempts at entering the same thermal repeatedly \emph{in the real world}, a luxury it doesn't have. On the other hand, thermalling controllers tested live \cite{allen-aiaa05,edwards-aiaa2008} rely on simple, fixed strategies that, unlike RL-based ones, don't take exploratory steps to gather information about a thermal. They were tested at high altitudes, where thermals are quite stable. However, below 200 meters, near the ground, thermals' irregular shape makes the lack of exploration a notable drawback, as we show in this paper. 

The main contribution of our work is framing and solving autonomous thermalling as partially observable Markov decision process (POMDP). A POMDP agent maintains a \emph{belief} about possible world models (in our case --- thermal models) and can explicitly predict how new information could affect its beliefs. This effectively allows the autopilot to build a simulator for a specific thermal in real time ``on the fly'', and trade off information gathering to refine it versus exploiting the already available knowledge to gain height. We propose a fast approximate algorithm tailored to this scenario that runs in real time on Pixhawk, a common autopilot hardware that has a 32-bit ARM processor with only 168MHz clock speed and 256KB RAM. On \suav{s} with a more powerful companion computer such as Raspberry Pi 3 onboard, our approach allows for thermalling policy generation with a full-fledged POMDP solver. Alternatively, our setup can be viewed and solved as a model-based Bayesian reinforcement learning problem \cite{ghavamzadeh-ftml15}.

For evaluation, we added the proposed algorithm to ArduPlane \cite{arduplane}, an open-source drone autopilot, and conducted a live comparison against ArduSoar, ArduPlane's existing soaring controller. This experiment comprised 14 missions, in which two RC sailplanes running each of the two thermalling algorithms onboard flew \emph{simultaneously} in weak turbulent windy thermals at altitudes below 200 meters. Our controller significantly outperformed ArduSoar in flight duration in 11 flights out of 14, showing that its unconventional thermalling trajectories let it take advantage of the slightest updrafts even when running on very low-power hardware.

%% file: background.tex
\section{Background}

\noindent
\textbf{Thermals and sailplanes.} \emph{Thermals} are rising plumes of air that originate above areas of the ground that give up previously accumulated heat during certain parts of the day (Figure \ref{fig:thermals}). They tend to occur in at least partly sunny weather several hours after sunrise above darker-colored terrain such as fields or roads and above buildings, but occasionally also appear where there are no obvious features on the Earth's surface. As the warm air rises, it coalesces into a ``column'', cools off with altitude and eventually starts sinking around the column's fringes. As any conceptual model, this representation of a thermal is idealized. In reality, thermals can be turbulent and irregularly shaped, especially at altitudes up to 300m.

\begin{figure}[h]
\begin{subfigure}{0.5\textwidth}
\includegraphics[width=0.5\textwidth]{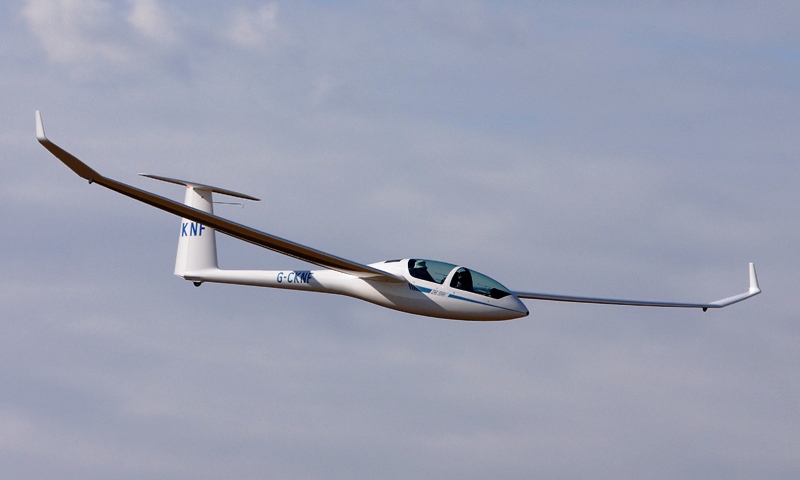}
\label{fig:sailplane}
\end{subfigure}%
~
\hspace{-4.5cm}
\begin{subfigure}{0.5\textwidth}
\includegraphics[width=0.44\textwidth]{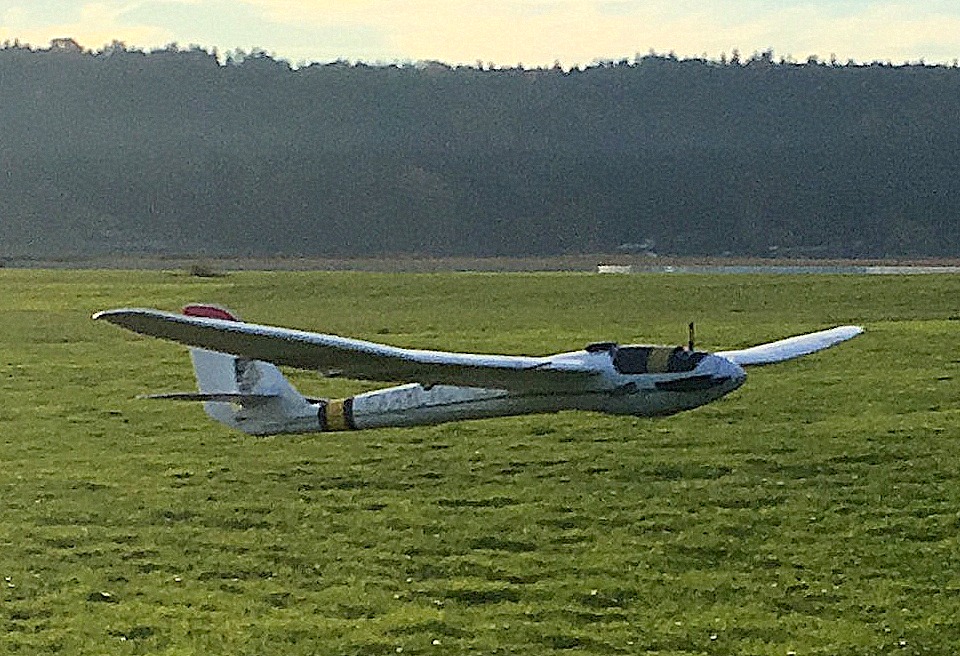}
\label{fig:radian}
\end{subfigure}
\caption{A full-size DG1000 sailplane (left, \emph{photo credit: Paul Hailday}) and Radian Pro remote-controllable model sailplane that serves as our \suav{} platform (right).}
\label{fig:sailplanes}
\end{figure}

Thermal updrafts are used by birds \cite{akos-bb10} and by human pilots flying \emph{sailplanes}. Sailplanes  (Figure \ref{fig:sailplanes}, left), colloquially also called \emph{gliders}, are a type of fixed-winged aircraft optimized for unpowered flight, although some do have a limited-run motor. To test our algorithms, we use a Radian Pro sailplane \suav{} (Figure \ref{fig:sailplanes}, right) controllable by a human from the ground.

Thermalling strategies depend on the distribution of lift within a thermal. Much of autonomous thermalling literature, as well as human pilots' intuition, relies on the bell-shaped model of lift \emph{in the horizontal cross-section of a thermal at a given altitude} \cite{wharington-icas98}. It assumes thermals to be approximately round, with vertical air velocity $w(x,y)$ being largest near the center and monotonically getting smaller towards the fringes:

\begin{equation}
w(x,y) = W_0 e^{-\frac{(x-x^{th})^2 + (y-y^{th})^2}{R_0^2}}
\label{eq:thermal}
\end{equation}

\noindent
Here, $(x^{th},y^{th})$ is the position of the thermal center \emph{at a given altitude}, $W_0$ is vertical air velocity, in m/s, at the center, and $R_0$ can be interpreted as the thermal's radius (Figure \ref{fig:thermals}, in red). Note that a thermal's lift doesn't disappear entirely more that $R_0$ meters from its center. In spite of its simplicity, we use this model in our controller for its low computational cost.\\
\backupsmall

\noindent
\textbf{MDPs and POMDPs.} Settings where an agent has to optimize its course of action are modeled as Markov Decision Processes (MDP). An MDP is a tuple $\langle \mathcal{S}, \mathcal{A}, \mathcal{T}, \mathcal{R}, s_0 \rangle$ where $\mathcal{S}$ is the set of possible joint agent/environment states, $\mathcal{A}$ is the set of actions available to the agent, $\mathcal{T}:\mathcal{S} \times \mathcal{A} \times \mathcal{S} \rightarrow [0,1]$ is a \emph{transition function} specifying the probability that executing action $a$ in state $s$ will change the state to $s'$, $\mathcal{R}:\mathcal{S} \times \mathcal{A} \times \mathcal{S} \rightarrow \mathbb{R}$ is a \emph{reward function} specifying the agent's reward for such a transition, and $s_0$ is the start state. An MDP agent is assumed to know the current state exactly. An optimal MDP solution is a mapping $\pi: \mathcal{S} \rightarrow \mathcal{A}$ called \emph{policy} that dominates all other policies under the \emph{expected reward value starting at $s_0$}:
\backupsmall

\begin{equation}
V^{\pi}(s_0) = E_{\mathcal{T}^{\pi}}\left[\sum^{\infty}_{i=0} \gamma^{i}\mathcal{R}(S_i, A^{\pi}_i, S_{i+1}) \biggr\rvert S_0 = s_0\right]
\label{eq:discr}
\end{equation}

\noindent
Here, $S_i$ and $A^{\pi}_i$ are random variables for the agent's state $i$ steps into the future and the action chosen by $\pi$ in that state, under trajectory distribution $\mathcal{T}^{\pi}$ induced by $\pi$ from $s_0$. \\

If the agent doesn't have full state knowledge but has access to noisy state observations (as in this paper's setting), it is in a \emph{partially observable} MDP (POMDP) setting \cite{astrom-jmaa65}. A POMDP is a tuple $\langle \mathcal{S}, \mathcal{A}, \mathcal{T}, \mathcal{R}, \mathcal{O}, \mathcal{Z}, b_0 \rangle$, where $\mathcal{S}, \mathcal{A}, \mathcal{T}$, and $\mathcal{R}$ are as in the MDP definition, $\mathcal{O}$ is the observation space of possible clues about the true state, and $\mathcal{Z}: \mathcal{A} \times \mathcal{S} \times \mathcal{O} \rightarrow [0,1]$ describes the probabilities of these observations for different states $s'$ where the agent may end up after executing action $a$. Since a POMDP agent doesn't know the world state, it maintains a \emph{belief} --- a state probability distribution given the observations received so far. Initially, this distribution is a just a prior $b_0:\mathcal{S} \rightarrow [0,1]$; the agent updates it using the Bayes rule. POMDP policies are belief-based and have the form $\pi: \mathcal{B} \rightarrow \mathcal{A}$, where $\mathcal{B}$ is the belief space. The optimization criterion amounts to finding a policy that maximizes
\backupsmall

\begin{equation}
V^\pi(b_0) = \sum_s b_0(s) V^{\pi}(s)
\end{equation}
\backupsmall

\noindent
where $V^{\pi}(s)$ comes from Equation \ref{eq:discr}. For a known initial belief state $b_0$, general POMDPs can be solved with different degrees of optimality using methods from the point-based family \cite{pineau-ijcai03} or variants of the POMCP algorithm \cite{silver-nips10}. \\
\backupsmall

\noindent
\textbf{Extended Kalman filter.} Since a POMDP agent's action choice depends on its belief, efficient belief updates are crucial. For Gaussian beliefs, the Bayesian update can be performed very fast using various Kalman filters. In the POMDP notation, under stationary transition and observation functions, the original Kalman filter (KF) \cite{kalman-jbe60} assumes $\mathcal{T}(s, a, s') = \mathcal{N}(s'|f(s,a), Q)$ and $\mathcal{Z}(a, s', o) = \mathcal{N}(o|h(a,s'), R)$, where $\mathcal{N}$ denotes a Gaussian, $Q$ and $R$ are \emph{process} and \emph{observation noise} covariance matrices, and transition and observation transformations $f(s,a)$ and $h(a,s')$ are linear in state and action features. If beliefs are Gaussian but$f(s,a)$ and $h(s',a)$ are non-linear, as in our scenario, they can be linearized locally, giving rise to the \emph{extended Kalman filter (EKF)}. 	

Concretely, suppose that for belief $b=\mathcal{N}(.|\vec{\mu}, \Sigma)$, the agent takes action $a$, gets observation $o$, and wants to compute  its new belief $b'=\mathcal{N}(.|\vec{\mu}', \Sigma')$. First, we linearize the transition $f(s,a)$ about the current belief by computing the Jacobian
\backupsmall
\begin{equation}
F = \frac{\partial f}{\partial \vec{\mu}} \biggr\rvert_{\vec{\mu}} 
\label{eq:ekf_first}
\end{equation}

\noindent
\noindent
and a new belief estimate not yet adjusted for observation $o$:

\backupsmall
\begin{equation}
\vec{\underline\mu}' = f(\vec{\mu}, a)\ \ \ \ \ \ \ \underline{\Sigma}' = F \Sigma F^\intercal + Q 
\end{equation}
\backupsmall

\noindent
Next, we linearize $h(a, s')$ around this uncorrected belief by producing the Jacobian

\begin{equation}
H = \frac{\partial h}{\partial \vec{\mu}} \biggr\rvert_{a,\vec{\underline{\mu}}'}
\end{equation}

\noindent
and compute the approximate Kalman gain

\begin{equation}
K = \underline{\Sigma}'H^\intercal (H \underline{\Sigma}'H^\intercal +R)^{-1}
\end{equation}

\noindent
Finally, we correct our intermediate belief for observations:

\backupsmall
\begin{equation}
\vec{\mu}' = \vec{\underline\mu}' + K(\vec{o} - h(a,\vec{\underline{\mu}}'))\ \ \ \ \ \ \ \ \ \Sigma' = (I-KH)\underline{\Sigma}'
\label{eq:ekf_last} \\
\end{equation}

\noindent
There are other filters for belief tracking in systems with non-linear dynamics, including unscented Kalman filter (UKF) \cite{julier-aero97} and particle filter (PF) \cite{delmoral-mdp96}. We choose EKF for its balance of approximation quality and low computational requirements.\\
\backupsmall

%% file: problem_formalization.tex
\section{Autonomous Thermalling as a POMDP}

Our formalization of autonomous thermalling is based on the  fact that within a thermal, a \suav{'s} probability of gaining altitude and altitude gain itself depend on the lift distribution (in our case, given by Equation \ref{eq:thermal}). Thus, in the MDP/POMDP terminology, the lift distribution model determines the transition and reward functions $\mathcal{T}$ and $\mathcal{R}$. Since Equation \ref{eq:thermal}'s parameters are initially unknown, neither are $\mathcal{T}$ and $\mathcal{R}$. However, the autopilot can keep track of a belief $b$ over model parameters $x^{th}, y^{th}, W_0$, and $R_0$ as it is receiving a stream of variometer data, making POMDP a natural choice for this scenario.

The belief $b$ characterizes the autopilot's uncertainty about the thermal's position, size, and strength. Based on its current belief, at any time the autopilot can \emph{simulate} observations it \emph{might} get from measuring lift strength at different locations, and estimate how this knowledge is expected to reduce its thermal model uncertainty. This simulation can also help assess the expected net altitude gain from following various trajectories. Thus, the autopilot can \emph{deliberately} select trajectories that reduce its model uncertainty in addition to providing lift. The ability to explicitly plan environment exploration is missing both from all thermalling controllers proposed so far, including those that use RL (see the Related Work section). As hypothesized by \citet{allen-2007}, \citet{edwards-aiaa2008}, and \citet{tabor-arxiv18}, and as demonstrated in our experiments, this ability is important, especially for thermalling in difficult conditions.  

The POMDP we formulate models the decision-making process \emph{once the sailplane is in a thermal}. Note that gaining altitude with the help of an updraft also involves thermal detection and timely exit. These are beyond the paper's scope; we rely on \citet{tabor-arxiv18}'s mechanisms to implement them. 

\begin{center}
\textbf{Practicalities, assumptions, and mode of operation}
\end{center}

First we identify our scenario's aspects that may violate POMDPs' assumptions, and adapt POMDPs to them. \\
\backupsmall

\noindent
\textbf{Thermalling as model predictive control.} POMDPs assume the environment to be stationary, i.e., governed by unchanging $\mathcal{T}$ and $\mathcal{R}$. Viewing thermalling in this way would require modeling thermal evolution, which is computationally expensive, error-prone, and itself involves assumptions. Instead, as is common for such scenarios, we view it as a model predictive control (MPC) problem \cite{garcia-automatica89}. Our controller gets invoked with a fixed frequency of $>1$Hz, solves the POMDP below approximately to choose an action for the current belief, and the process repeats at the next time step. \\
\backupsmall

\noindent
\textbf{Modelling assumptions.} The policy quality of our controller depends on the validity of the following assumptions:

\begin{itemize}
\item{\textbf{Assumption 1.}} \emph{The thermal changes with time no faster than on the order of tens of seconds, and is approximately the same within a few meters of vertical distance.} This ensures that the world model used by the controller to choose the next action isn't too different from the world where this action will be executed. Recall that Equation \ref{eq:thermal} applies to a given altitude, and as the \suav{} thermals, its altitude changes. The assumption is that, locally, the thermal doesn't change with altitude too much.

\item{\textbf{Assumption 2.}} \emph{The thermal doesn't move w.r.t. the surrounding air.} The air mass containing both the thermal and the \suav{} may move w.r.t. the ground due to \emph{wind}. We effectively assume that the thermal moves at the wind velocity. While this assumption isn't strictly true for thermals \cite{telford-jas70}, it is common in the autonomous thermalling literature, and its violation doesn't seriously affect our controller, which recomputes the policy frequently.

\item{\textbf{Assumption 3.}} \emph{The \suav{} is flying at a constant airspeed, and the thermalling controller can't change the \suav{'s} pitch angle at will.} During thermalling, pitch angle control is necessary only for maintaining airspeed and executing coordinated turns, which can be done by the lower levels of the autopilot. If the thermalling controller could change pitch directly, it might attempt to ``cheat'' by using it to convert kinetic energy into potential to gain altitude, instead of exploiting thermal lift to do so.

\item{\textbf{Assumption 4.}} \emph{The thermal has no effect on the \suav{'s} horizontal displacement.} This holds for the model in Equation 1 because that model disregards turbulence.
\end{itemize}

\noindent
\textbf{A note on reference frames.} Wind complicates some computations related to \suav{'s} real-world state, because the locations of important observations such as lift strength are given in Earth's reference frame, in terms of GPS coordinates, whereas the \suav{} drifts with the wind. The autopilot maintains a wind vector estimate and uses it to translate GPS locations to the air mass's reference frame \cite{hazard-aiaa10,tabor-arxiv18}.

To make wind correction unnecessary, our POMDP model reasons entirely in the reference frame of the air mass. Due to Assumption 2 above, in this frame \suav{'s} air velocity fully accounts for the thermal's displacement w.r.t. the \suav. When solving the POMDP involves simulating observations, their locations are generated directly in this reference frame too.

\begin{center}
\textbf{POMDP Formulation}
\end{center}
\vspace{-0.05in}

\noindent
The components of the thermalling POMDP are as follows:\\
\backupsmall

\noindent
\textbf{State space $\mathcal{S}$} consists of vectors $(\uavstate, \thmlstate)$ describing the joint state of the \suav{} ($\uavstate$) \emph{and} the thermal ($\thmlstate$). In particular, $\uavstate = (\uavpos, \airspeed, \yaw, \roll, \rollrate, \alt)$ and $\thmlstate = (\thmlpos, \thmlz, \thmlr)$, where \uavpos gives the 2-D location, in meters, of the \suav{} w.r.t. an arbitrary but fixed origin of the air mass coordinate system, \airspeed is \suav{'s} airspeed in m/s, \yaw is its heading -- the angle between north and the direction in which it is flying, \roll is its bank angle, \rollrate is its rate of roll, and \alt is its altitude w.r.t. the mean sea level (MSL). The thermal state vector consists of \thmlpos $= (x^{th}, y^{th})$ -- the 2-D position of the thermal model center \emph{relative to the \suav{} position}, \thmlz\ -- the vertical lift in m/s at the thermal center and \thmlr\ -- the thermal radius (Figure \ref{fig:thermals}).

Wind vector $\vec{w}$ and pitch angle are notably missing from the state space definition due to Assumptions 2 and 3. \\
\backupsmall

\noindent
\textbf{Action space $\mathcal{A}$} is a set of arc-like \suav{} trajectory segments originating at the \suav{'s} current position, parametrized by duration $T_{\mathcal{A}}$ common to all actions and indexed by a set of bank angles $\{\roll_{1}, \ldots, \roll_{n}\}$. Each trajectory corresponds to a coordinated turn at bank angle $\roll_{i}$ for $T_{\mathcal{A}}$ seconds. It is not exactly an arc, because attaining bank angle $\roll_{i}$ takes time dependent on \suav{'s} roll rate $\rollrate$ and current bank angle $\roll$.\\
\backupsmall

\noindent
\textbf{Transition function $\mathcal{T}$} gives the probability of the joint system state $s'$ after executing action $a$ in state $(s^{u}, s^{th})$. Under Assumptions 2 and 4, $\mathcal{T}$ is described by the \suav{'s} dynamics and Gaussian process noise: 
\backupsmall

\begin{equation}
\mathcal{T}(s,a,s') = \mathcal{N}(s'|f_{T_{\mathcal{A}}}(s,a), Q)
\end{equation}

\noindent
Here, $f_{T_{\mathcal{A}}}$ captures the \suav{'s} dynamics, $Q$ is the process noise covariance matrix, and $\mathcal{T}$ satisfies a crucial property: \emph{$\mathcal{T}$ does not modify thermal parameters $W_0$, $R_0$, and thermal center $\vec{p}^{th}$ w.r.t. \suav{'s} position $\vec{p}^{u}$ in $s$.} Intuitively, thermal doesn't change just because the \suav{} moves. The change in the thermal center position is purely due to \suav{'s} motion and the fact that the thermal's position is relative to \suav{'s}. \\
\backupsmall

\noindent
\textbf{Reward function $\mathcal{R}$} for states $s$ and $s'$ is $(h_s - h_{s'})$, the resulting change in altitude. This definition relies on Assumption 3. Without it, thermalling controller could gain altitude by manipulating pitch, not by exploiting thermal lift.\\
\backupsmall

\noindent
\textbf{Observation set $\mathcal{O}$} consists of the possible sets of sensor readings of airspeed sensor, GPS, and barometer readings that the \suav{} might receive while executing an action $a$. \\
\backupsmall

\noindent
\textbf{Observation function $\mathcal{Z}$} assigning probabilies $\mathcal{Z}(a, s', o)$ to various observation sets $o$ is governed by action $a$'s trajectory and Gaussian process noise covariance matrix $R$. Each of \suav{'s} sensors operates at some frequency $\xi$. During $a$'s execution, this sensor generates a sequence of datapoints of length $\xi T_{\mathcal{A}}$, each sampled from a 0-mean Gaussian noise around the corresponding intermediate point on $a$'s trajectory. The probability of a set of sensor data sequences produced during $a$'s execution is the product of corresponding Gaussians.\\
\backupsmall

\noindent
\textbf{Initial belief $b_0$} is given by a product of two Gaussians:
\begin{equation*}
b_0(s) = b_0(s^u)b_0(s^{th}) =  \mathcal{N}(s^u|s^{u}_0, \Sigma_0^u)\mathcal{N}(s^{th}|s^{th}_0, \Sigma_0^{th}),
\end{equation*}

\noindent
where $s^{u}_0$ has $\vec{p}_0^u = (0,0)$ for mathematical simplicity, because \suav{'s} position is w.r.t. the air mass and for POMDP's purposes isn't tied to any absolute location, and $s^{u}_0$'s other components are \suav{'s} current airspeed, roll, etc. $s^{th}_0$ is initialized with estimates of thermals generally encountered in the area of operations. $\Sigma_0^u$ and $\Sigma_0^{th}$ are diagonal matrices.

Keeping the belief components for \suav{} and thermal state separate is convenient for computational reasons. The \suav{} state observations are heavily filtered sensor signals, so in practice \suav{'s} state can be treated as known. At the same time, thermal state belief, especially the initial belief $b^{th}_0$, has a lot of uncertainty and will benefit from a full belief update. \\
\backupsmall

\noindent
\textbf{Belief updates} are a central factor determining the computational cost of solving a POMDP. Our primary motivation for defining the initial belief $b_0$ to be Gaussian, in spite of possible inaccuracies, is that we can repeatedly use the EKF defined in Equations \ref{eq:ekf_first}-\ref{eq:ekf_last} to update $b_0$ with new observations and get a Gaussian posterior $b'$ at every step:
\vspace{-0.1cm}
\begin{equation}
b' = \mbox{EKF\_update}(b_0, a, \vec{o})
\label{eq:ekf_pomdp}
\end{equation}
\noindent
The $\mbox{EKF\_update}$ routine implicitly uses transition and observation functions $\mathcal{T}$ and $\mathcal{Z}$ from the POMDP's definition.

%% file: algorithm.tex
\section{The Algorithm}

\begin{figure}[!t]
 \removelatexerror
\begin{algorithm}[H]
\label{a:tln}
\caption{\ouralgo}
Input: POMDP $\langle {\cal S}, {\cal A}, {\cal T}, {\cal R}, {\cal O}, {\cal Z}, b_0 \rangle$, $N$ -- number of thermal state belief samples, $T^{\mbox{exploit}}_{\mathcal{A}}$ -- planning horizon for exploitation mode, $T^{\mbox{explore}}_{\mathcal{A}}$ -- planning horizon for exploration mode, $\Delta t$ -- action trajectory time resolution, THERMAL\_CONFIDENCE\_THRES -- thermal state confidence threshold \label{l:input}\\

$b_0 = \mathcal{N}(s^{u}_0, \Sigma_0^u)\mathcal{N}(s^{th}_0, \Sigma_0^{th}) \leftarrow$ current belief \\
\ \\
\uIf{$tr(\Sigma_0^{th}) < \mbox{THERMAL\_CONFIDENCE\_THRES}$}{\label{l:check}
Go to \underline{EXPLOIT}
}\Else{
Go to \underline{EXPLORE}
}
\ \\
\underline{EXPLORE} \Begin{
\ForEach{$a \in  \mathcal{A}$}
{
$\mbox{Traj}_{a} \leftarrow \mbox{\underline{SimActionTraj}}(s^{u}_0, \roll_a, \Delta t, T^{\mbox{explore}}_{\mathcal{A}})$ \label{l:suavstate_lore}\\
\ForEach{$i=1 \ldots N$}{
$s^{th}_i \leftarrow \mbox{\underline{SampleThermalState}}(b_0)$ \\
\ForEach{$t=1 \ldots \mbox{Traj}_{a}.\mbox{Length}$}{
$\vec{o}_{a,i,t} \leftarrow \mbox{\underline{SimObservation}}(\mbox{Traj}_{a}[t], s^{th}_i)$ \\
$b_{a,i,t} \leftarrow \mbox{EKF\_update}(b_{a,i,t-1},a_{[t-1,t]},\vec{o}_{a,i,t})$ \label{l:ekf}\\
}
$\Sigma^{th}_{a,i} \leftarrow \mbox{covariance of } b_{a,i,Traj_{a}.\mbox{Length}}$ \\
}
$\mbox{Uncertainty}_a \leftarrow \frac{\sum_{i=1}^N tr(\Sigma^{th}_{a,i})}{N}$ \label{l:unc}\\
}
$a^* \leftarrow \mbox{argmin}_{a \in \mathcal{A}} \mbox{Uncertainty}_a$ \\
Return $a^*$ \\
}
\ \\
\underline{EXPLOIT} \Begin{
\ForEach{$a \in  \mathcal{A}$}
{
$\mbox{Traj}_{a} \leftarrow \mbox{\underline{SimActionTraj}}(s^{u}_0, \roll_a, \Delta t, T^{\mbox{exploit}}_{\mathcal{A}})$ \label{l:suavstate_loit} \\
\ForEach{$i=1 \ldots N$}{
$s^{th}_i \leftarrow \mbox{\underline{SampleThermalState}}(b_0)$ \\
\ForEach{$t=1 \ldots \mbox{Traj}_{a}.\mbox{Length}$}{
$w_{a,i,t} \leftarrow \mbox{\underline{SimLift}}(\mbox{Traj}_{a}[t], s^{th}_i)$ \label{l:lift}\\
}
$\mbox{AltGain}_{a,i} \leftarrow \sum^{\mbox{Traj}_{a}.\mbox{Length}}_{t = 1} w_{a,i,t} \Delta t$ \label{l:altgain}\\
}
$\mbox{ExpAltGain}_a \leftarrow \frac{\sum_{i=1}^N \mbox{AltGain}_{a,i}}{N}$ \\
}
$a^* \leftarrow \mbox{argmax}_{a \in \mathcal{A}} \mbox{ExpAltGain}_a$ \\
Return $a^*$ \\
}
\end{algorithm}
\vspace{-0.3in}
\end{figure}

Although using EKF-based belief updates reduces the computational cost of solving a POMDP, solving an EKF-based POMDP near-optimally on common low-power flight controllers such as APM and Pixhawk is still much too expensive. The algorithm we present, \ouralgo\ (Algorithm \ref{a:tln}), makes several fairly drastic approximations in order to be solve the POMDP model. While they undoubtedly lead to sacrifices in solution quality, \ouralgo\ still retains a controller's ability to explore in a (myopically) guided way, which, we claim, gives it advantage in messy thermals at low altitudes. This section presents a high-level description of \ouralgo, while further details about its tradeoffs and implementation are contained in the \extradata{}. 

To further reduce belief update cost, \ouralgo\ always uses the belief mean as the estimate of the \suav{} state; see, e.g., line \ref{l:suavstate_lore} of Algorithm \ref{a:tln}. For the thermal part of the state, however, it uses the full belief over that part. For a Gaussian, it is natural to take covariance trace as a measure of uncertainty, and this is what \ouralgo\ does (lines \ref{l:check}, \ref{l:unc}). 

Conceptually, \ouralgo\ is split into two parts: exploration and exploitation. Whenever it is invoked, it first analyzes the covariance of the controller's current belief \emph{about the thermal state} (line \ref{l:check}). If its trace is below a given threshold --- an input parameter --- \ouralgo\ chooses an action aimed at exploiting the current belief. Otherwise, it tries to reduce belief uncertainty by doing exploration. In effect, \ouralgo\ switches between two approximations of the POMDP reward function $\mathcal{R}$, an exploration- and a lift-oriented one.

In either mode, \ouralgo\ performs similar calculations. Its action set is a set of arcs parametrized by discrete roll angles, e.g. -45, -30, -15, 0, 15, 30, and 45 degrees. For each of them it computes the sequence of points that would result from executing a coordinated turn at that roll angle for $T^{\mbox{explore}}_{\mathcal{A}}$ or $T^{\mbox{exploit}}_{\mathcal{A}}$ seconds (lines \ref{l:suavstate_lore}, \ref{l:suavstate_loit}). See the Implementation Details section for more information about this computation.

Then, \ouralgo\ samples $N$ thermal states from the current belief and for each of these ``imaginary thermals'' \emph{simulates} the reward of executing each of the above action trajectories in the presence of this thermal. In explore mode, action reward is the reduction in belief uncertainty, so at each of the trajectory points \ouralgo\ generates a hypothetical observation and uses it in a sequence of \emph{imaginary} belief updates (line \ref{l:ekf}). The trace of the last resulting belief measures thermal belief uncertainty after the action's hypothetical execution. For execution in the real world, \ouralgo\ chooses the action that minimizes this uncertainty.

In the exploit mode, at each generated point of each action trajectory, \ouralgo\ measures hypothetical lift (line \ref{l:lift}) instead of generating observations, and then integrates the lift along the trajectory to estimate altitude gain (line \ref{l:altgain}). The action with the maximum altitude gain estimate "wins".

\begin{figure*}
\vspace{-0.1in}
\begin{subfigure}{0.50\textwidth}
\centering
\includegraphics[width=\linewidth]{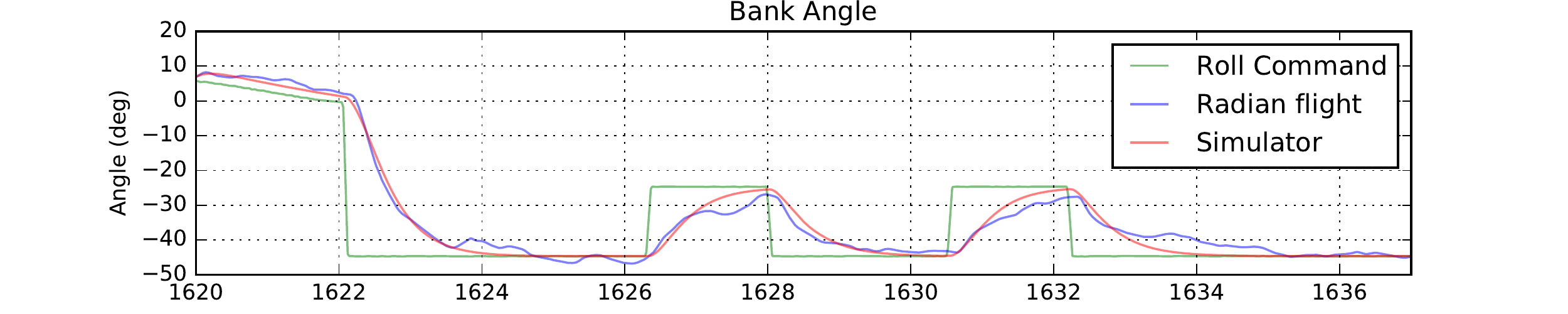}
\vspace{-0.5cm}
\caption{Bank angle}
\label{fig:bank_angle}
\end{subfigure}
\begin{subfigure}{0.50\textwidth}
\centering
\includegraphics[width=\linewidth]{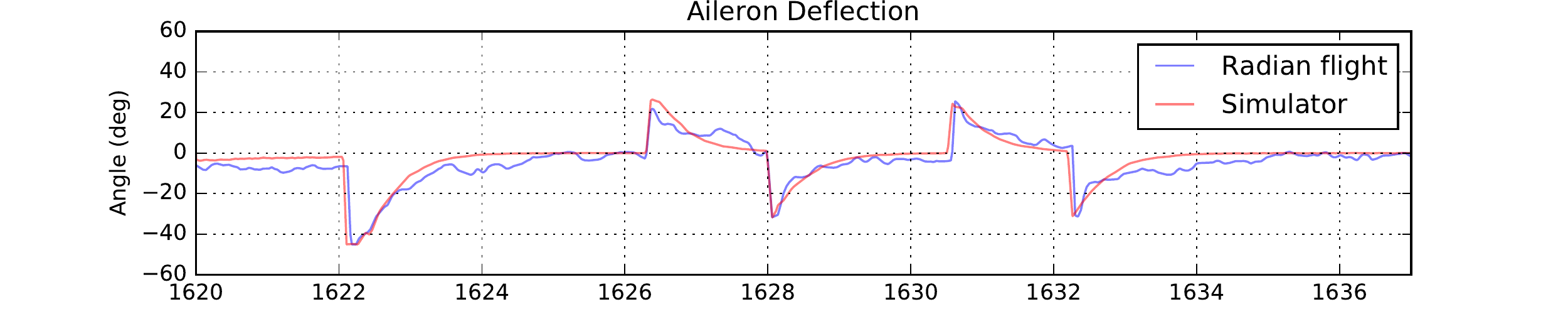}
\vspace{-0.5cm}
\caption{Aileron deflection angle}
\label{fig:defl_angle}
\end{subfigure}
\vspace{-0.05in}
\caption{\small Comparison between actual flight data and its predictions by the model used in \ouralgo\ for the same input roll commands.}
\label{fig:model_match}
\vspace{-0.25in}
\end{figure*}

\ouralgo's performance in practice critically relies on the match between the \emph{actual} turn trajectory (a sequence of 2-D locations) the \suav{} will follow for a commanded bank angle and the turn trajectory \emph{predicted by our controller} for this bank angle and used by \ouralgo\ during planning. Our controller's trajectory prediction model is in the \extradata{}. It uses several airframe type-specific parameters that had to be fitted to data gathered by entering the \suav{} into turns at various bank angles. The model achieves a difference of $< 1$m between predicted and actual trajectories. Figure \ref{fig:model_match} shows the match between the model-predicted and actual bank and aileron deflection angle evolutions for several roll commands.

Our \ouralgo\ implementation is available at \emph{\href{https://github.com/Microsoft/Frigatebird}{https://github.com/Microsoft/Frigatebird}}. It is in C++ and is based on ArduPlane 3.8.2, an open-source autopilot for fixed-wing drones \cite{arduplane} that has \citet{tabor-arxiv18}'s ArduSoar controller built in. We reused the parts of it that were conceptually common between ArduSoar and \ouralgo, including the thermal tracking EKF and thermal entry/exit logic. ArduSoar and \ouralgo\ implementations differ only in how they use the EKF to choose the \suav{'s} next action. Running \ouralgo\ onboard a \suav{} requires picking input parameters (Algorithm \ref{a:tln}, line \ref{l:input}) that result in reasonable-quality solutions within the short ($< 1$s) time between controller invocations on the flight controller hardware. All parameters values from our experiments are available in a .param file together with \ouralgo's implementation.

%% file: related_work.tex
\section{Related Work}

While our approach can be regarded as solving an POMDP, it is also an instance of Bayesian reinforcement learning (BRL) \cite{filatov-cta00,ghavamzadeh-ftml15}. In BRL, the agent starts with a prior distribution over model dynamics $\mathcal{T}$ and $\mathcal{R}$, and maintains a posterior over them as it gets observations. Viewing $\mathcal{T}$ and $\mathcal{R}$ as part of the augmented state space $\mathcal{S}' = \mathcal{S} \times \mathcal{T} \times \mathcal{R}$ reveals that BRL is equivalent to a special POMDP called Bayes-adaptive MDP (BAMDP) \cite{duff-2002}, which \ouralgo\ ends up solving. Nonetheless, due to the computational constraints of our platform, \ouralgo\ is very different from existing POMDP and BRL solvers such as POMCP \cite{silver-nips10} and BAMCP \cite{guez-jair13}. However, BRL-induced POMDPs with belief tracking using EKF were studied by \citet{slade-iros17}, and the idea of separating uncertainty reduction from knowledge exploitation for approximate BRL is similar to \citet{dearden-uai99}.

There is vast literature on automatically exploiting thermals \cite{wharington-1998,wharington-icas98,allen-2007,edwards-aiaa2008,hazard-aiaa10,reddy-pnas16,lecarpentier-jfpda17}, orographic lift \cite{langelaan-aiaa07,fisher-bb16}, wind gusts \cite{langelaan-aiaa08}, wind fields \cite{lawrance-icra11}, and wind gradients \cite{lawrance-icra09,bird-aiaa14}, as well as on planning flight paths to extend fixed-wing \suav{} endurance and range \cite{edwards-ja10,lawrance-icra11,chakrabarty-jgcd11,chung-jrr15,lecarpentier-jfpda17,depenbusch-jfr17}. Soaring patterns have also been studied for birds \cite{akos-bb10}. Since our paper focuses exclusively on thermalling, here we survey papers from this subarea.

In an early autonomous thermalling work, Wharington et al \cite{wharington-1998,wharington-icas98} used RL \cite{sutton-98} in simulation to learn a thermalling strategy under several simplifying assumptions. \citet{reddy-pnas16} also used RL but removed some of \citet{wharington-icas98}'s simplifications. In particular, they built a far more detailed thermal model that reflects updrafts' messy, turbulent nature. However, neither of these techniques have been deployed on real UAVs so far, and how this could be done is conceptually non-obvious. These works' version of RL would require executing many trials \emph{in the real world} in order to learn a thermalling strategy for a given situation, including the ability to restart a trajectory at will --- a luxury unavailable to real-world \suav{s}. Given the wide variability in atmospheric conditions, it is also unclear whether this method can be successfully used to learn a policy offline for subsequent deployment onboard a \suav. Our POMDP model circumvents these challenges facing classical RL in the autonomous soaring scenario, since the former allows sampling trajectories during flight \emph{on the flight controller's hardware running a data-driven simulator built on the go}. However, incorporating \citet{reddy-pnas16}'s elaborate thermal model into a POMDP-based controller could yield even more potent thermalling strategies.  

\citet{allen-2007} were the first to conduct an extensive \emph{live} evaluation of an automatic thermalling technique. They employed a form of Reichmann rules \cite{allen-2007,reichmann-93} --- a set of heuristics employed for thermal centering by human pilots --- to collect data for learning thermal location and radius in flight. \citet{edwards-aiaa2008} mitigated this approach's potential issue with biased thermal parameter estimates. In a live flight test conducted primarily at altitudes over 700m, the resulting method along with inter-thermal speed optimization kept a sailplane airborne for 5.3 hours, the record for automatic soaring so far, with thermal centering running on a laptop on the ground. It is not clear whether it is possible to implement it as part of a fully autonomous autopilot on such a computationally constrained device as a Pixhawk, and how such an implementation, if realistic, would perform in less regular low-altitude thermalling conditions. Another controller based on Reichmann rules was introduced by \citet{andersson-jgd12}. \citet{daugherty-aiaa14} improved on it by using a Savitzky-Golai filter to estimate total specific energy and its derivatives, which reduced the estimation lag compared to \citet{andersson-jgd12} and allowed latching onto smaller thermals in simulation. \citet{andersson-jgd12}'s controller itself, like those of \citet{edwards-aiaa2008} and \citet{allen-2007}, was evaluated on a real \suav{} and did well at altitudes over 400m AGL, but relative live performance of all these controllers is unknown.

Last but not least, our live study compares \ouralgo\ against ArduSoar \cite{tabor-arxiv18}, the thermalling controller of a popular open-source autopilot for fixed-wing \suav{s} called ArduPlane \cite{arduplane}. ArduSoar's implementation of thermal-tracking EKF and thermal entry/exit logic is shared with \ouralgo. When ArduSoar detects a thermal, it starts circling at a fixed radius around the EKF's thermal position estimate mean. ArduSoar's thermal tracking was inspired by \citet{hazard-aiaa10}'s work, which used a UKF instead of EKF for this purpose and evaluated a number of fixed thermalling trajectories in simulation.

Thus, several aspects distinguish our work from prior art. On the theoretic level, we frame thermalling as a POMDP, which allows principled analysis and practical onboard solutions. We also conduct the first live side-by-side evaluation of two thermalling controllers. Last but not least, this evaluation is done in an easily replicable experimental setup.

%% file: experiments.tex
\section{Empirical Evaluation}

Our empirical evaluation aimed at comparing \ouralgo's performance as part of an end-to-end \suav{} autopilot to a publicly available thermalling controller for \suav{s}, ArduSoar, in challenging real-life thermalling conditions. We did this via side-by-side live testing of the two soaring controllers. The choice of ArduSoar as a baseline is motivated by several considerations. First, despite its simplicity, it performs well at higher altitudes. Second, it incorporates central ideas from several other strong works, including \citet{allen-2007,edwards-aiaa2008}, and \citet{hazard-aiaa10}. Third, the availability of source code, documentation, parameters, and hardware for it eliminates from our experiments potential performance differences due to hardware capabilities, implementation tricks, or mistuning.

A field study of a thermalling controller is complicated by exogenous factors such as weather, so before presenting the results we elaborate on the evaluation methodology.\\
\backupsmall

\begin{figure}[h]
\vspace{-0.15in}
\centering
\includegraphics[width=0.41\textwidth]{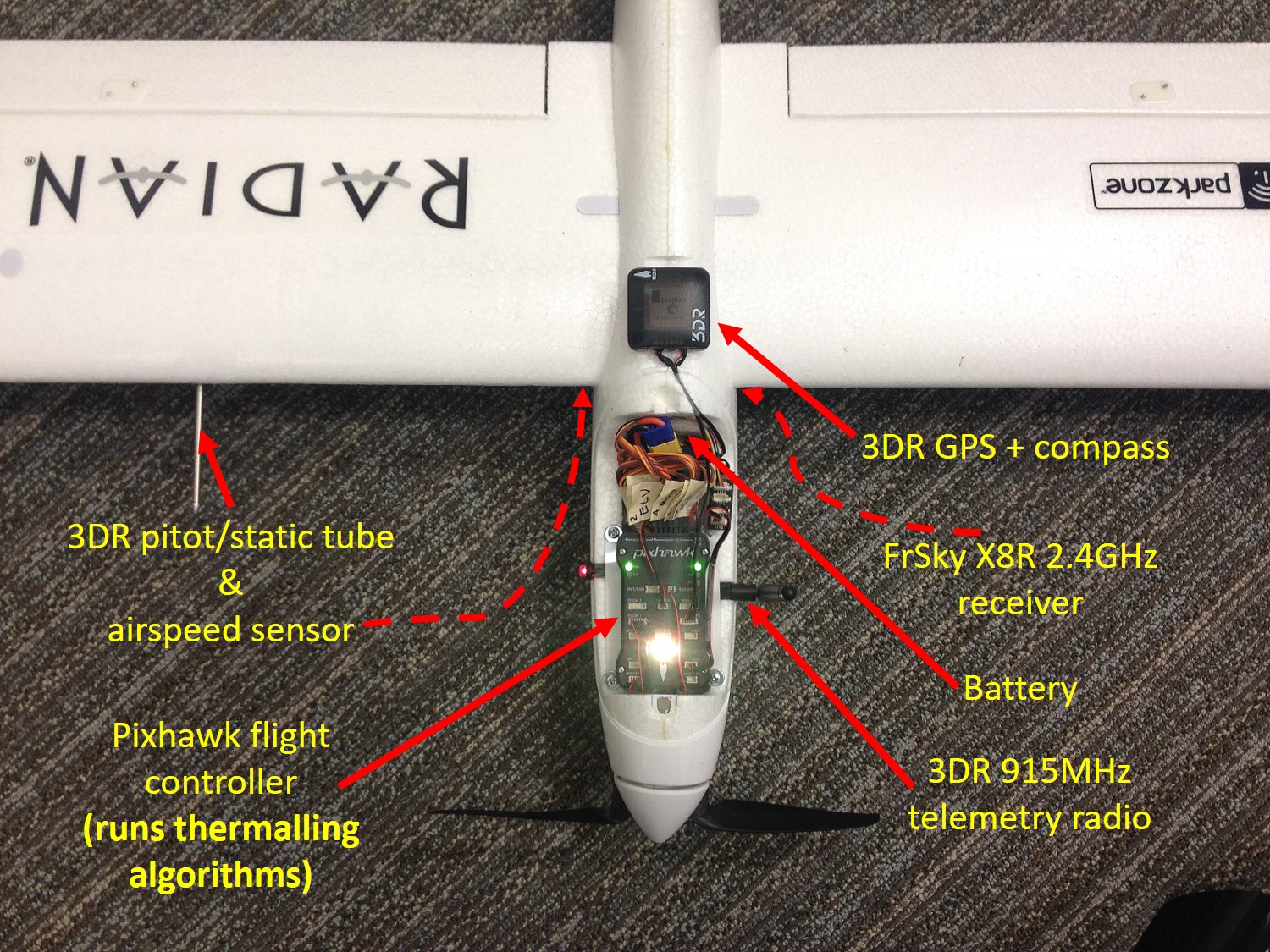}
\caption{Radian Pro \suav{'s} electronic equipment}
\label{fig:eq_diag}
\vspace{-0.15in}
\end{figure}

\noindent
\textbf{Equipment.} We used two identical off-the-shelf Radian Pro remote-controllable sailplanes as \suav{} airframes. They are made of styrofoam, have a 2m wingspan, 1.15m length, and carry a motor that can run for a limited duration. To enable them to fly autonomously, we reproduced the setup from \citet{tabor-arxiv18}, installing on each a 3DR Pixhawk flight controller (32-bit 168MHz ARM processor, 256KB RAM, 2MB flash memory), a GPS, and other peripherals, as shown in Figure \ref{fig:eq_diag}. A human operator could take over control at will using an X9D+ remote controller. All onboard electronics and the motor was powered by a single 3-cell 1300 mAh LiPo battery.\\
\backupsmall

\noindent
\textbf{Software.} Both Radian Pros ran the ArduPlane autopilot modified to include \ouralgo, as described in the Implementation Details section of the \extradata{}. During each flight, \ouralgo\ was enabled on one Radian Pro, and ArduSoar was enabled on the other. We used Mission Planner v1.3.49 \cite{mp} as ground control station (GCS) software to monitor flight telemetry in real time. ArduPlane was tuned separately on each airframe using a  standard procedure \cite{arduplane}. The parameters not affected by tuning were copied from \citet{tabor-arxiv18}'s, setup, including the target airspeed of 9 m/s. The airspeed sensors were recalibrated before every flight.\\
\backupsmall

\begin{figure*}
\vspace{-1.5cm}
\begin{subfigure}{0.5\textwidth}
\centering
\includegraphics[width=\textwidth]{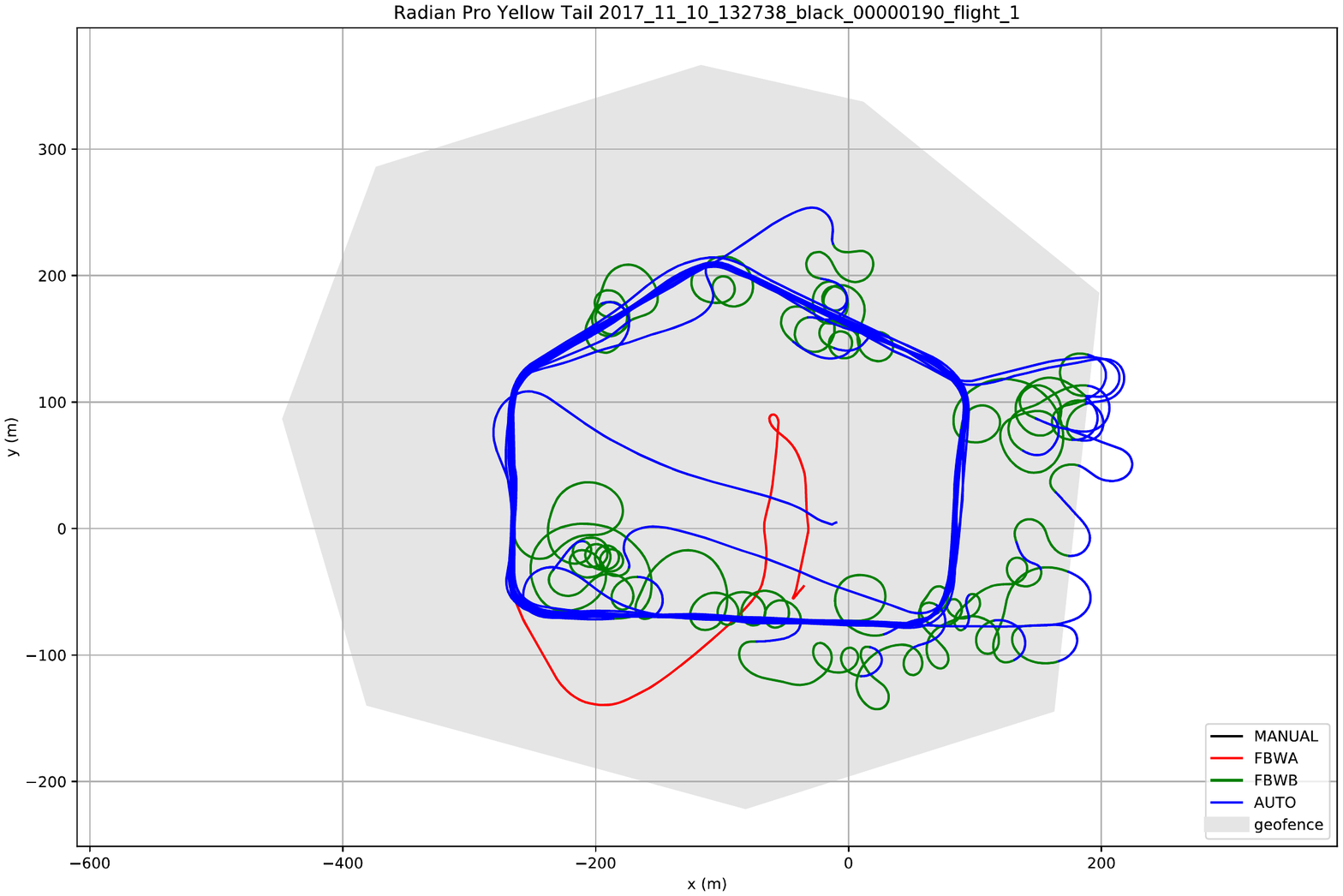}
\vspace{-1.5cm}
\caption{The Field test site and \ouralgo's typical flight path at it.}
\label{fig:pomdsoar_60acres}
\end{subfigure}
\begin{subfigure}{0.49\textwidth}
\centering
\vspace{0.1cm}
\includegraphics[width=\textwidth]{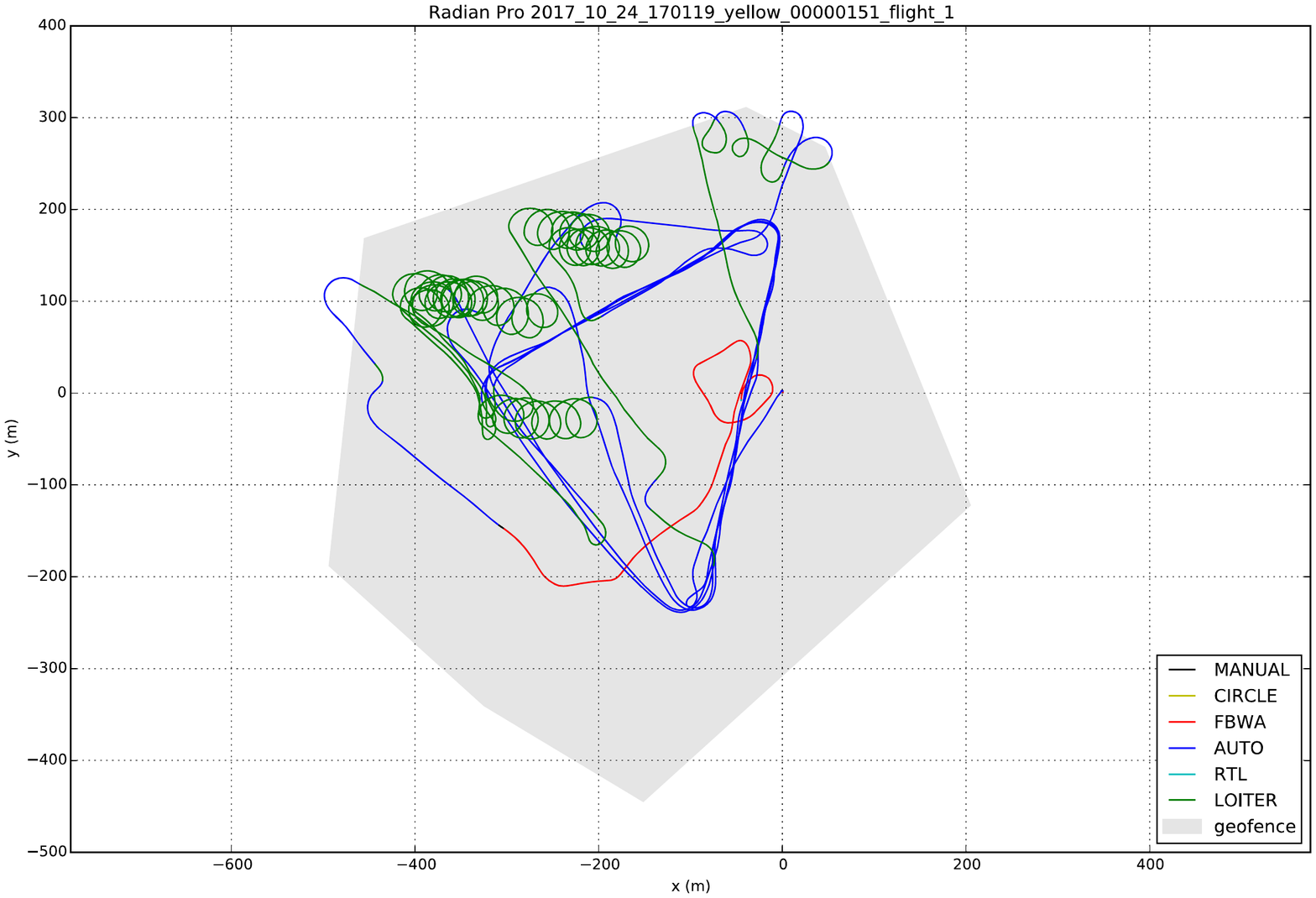}
\vspace{-1.5cm}
\caption{The Valley test site and ArduSoar's typical flight path at it.}
\label{fig:ardusoar_carnation}
\end{subfigure}
\caption{\small The Field and Valley test site layout and typical mission flight paths due to \ouralgo\ and ArduSoar. When not thermalling (AUTO mode), the \suav{s} follow a fixed sequence of waypoints, resulting in a pentagon-shaped path at the Field site and a triangle-shaped one at the Valley. During thermalling, the \suav{s} are allowed to deviate from this path anywhere within the geofenced region shown in grey. In the thermalling mode (FBWB for \ouralgo, LOITER for ArduSoar, green sections of the paths), \ouralgo\ does a lot of exloration, yielding irregularly-shaped meandering trajectories. ArduSoar, due to a more rigid thermalling policy, yields spiral-shaped paths that result in little exploration. Both controllers are forced to give up thermalling and switch to AUTO mode if the \suav{} breaches the geofence.}
\label{fig:test_sites}
\end{figure*}

\begin{figure*}
\vspace{-0.05in}
\hspace{0.2cm}
\begin{subfigure}{0.46\textwidth}
\centering
\includegraphics[width=\textwidth]{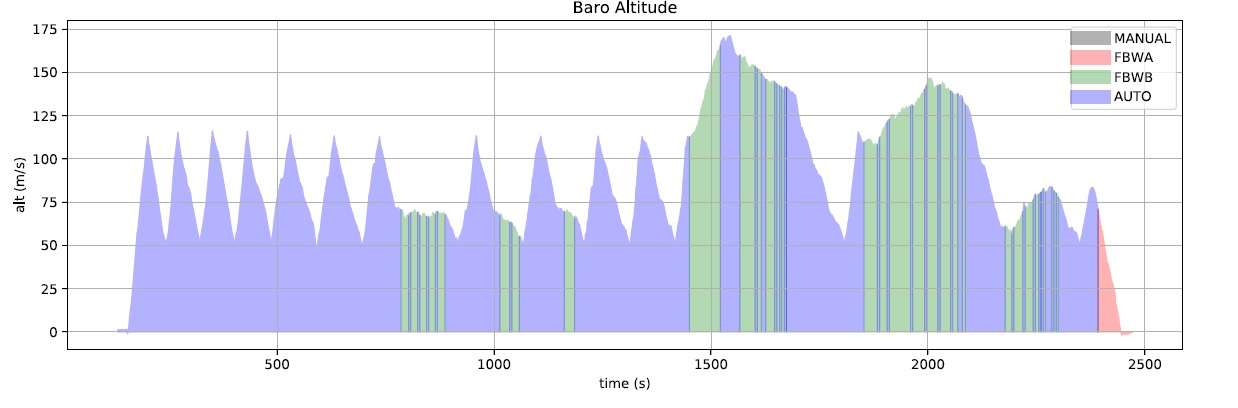}
\vspace{-0.7cm}
\caption{Altitude plot for a typical thermalling flight.}
\label{fig:thermalling_alt}
\end{subfigure}
\hspace{0.9cm}
\begin{subfigure}{0.43\textwidth}
\centering
\includegraphics[width=\textwidth]{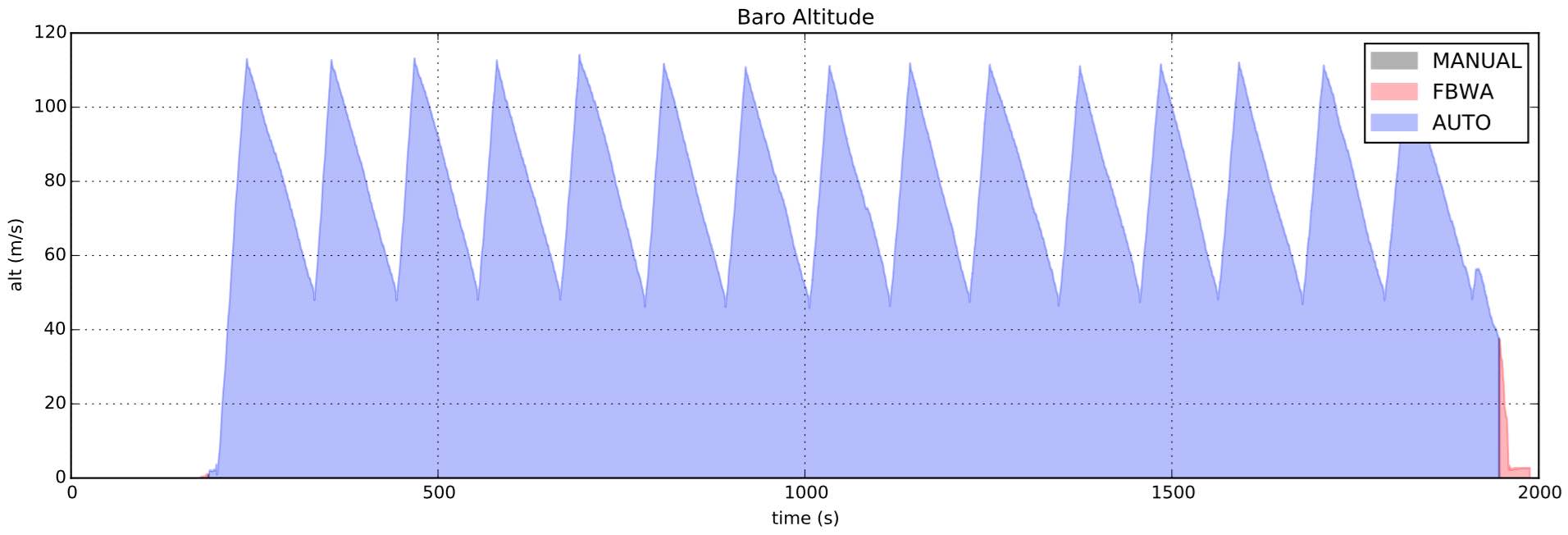}
\vspace{-0.6cm}
\caption{Altitude plot for a typical baseline measurement flight.}
\label{fig:baseline_alt}
\end{subfigure}
\vspace{-0.1cm}
\caption{\small Altitude vs. time plots for thermalling and baseline flights illustrating the mission profile. After being hand-tossed into the air in AUTO mode, each Radian automatically started its motor and climbed to the altitude given by SOAR\_ALT\_CUTOFF. At that altitude, it would shut down the motor and glide down, still in AUTO mode, until it either reached SOAR\_ALT\_MIN or detected a thermal. In the latter case, it would abandon waypoint following and enter a thermalling mode (FBWB for \ouralgo, LOITER for ArduSoar) until it gained altitude up to SOAR\_ALT\_MAX, descended to SOAR\_ALT\_MIN if the thermalling attempt was unsuccessful, or hit the geofence. In all these cases, the AUTO mode would engage automatically, forcing the \suav{} to give up thermalling, guiding it to the next waypoint, and turning on the motor if necessary to climb from SOAR\_ALT\_MIN to SOAR\_ALT\_CUTOFF. For Field and Valley test sites, SOAR\_ALT\_MIN is 50m and 30m, SOAR\_ALT\_CUTOFF is 110m and 130m, and SOAR\_ALT\_MAX is 160m and 180m, respectively.}
\label{fig:alt_profile}
\end{figure*}


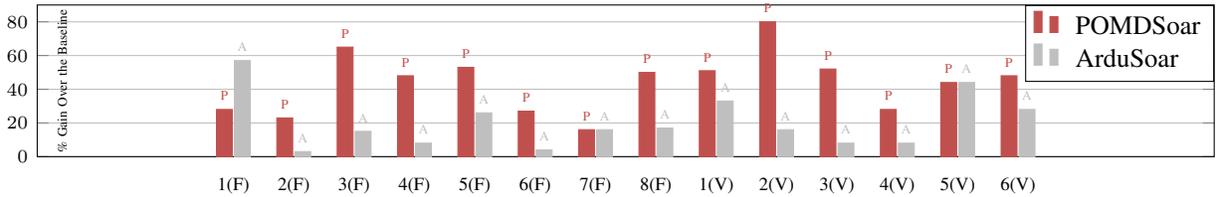
\begin{figure*}
\vspace{-0.13in}
\centering
\begin{tikzpicture}
    \begin{axis}[
        width  = 0.95 * \textwidth,
        height = 3.6cm,
        major x tick style = transparent,
        ybar=2*\pgflinewidth,
        bar width=6pt,
        ymajorgrids = true,
        ylabel = {\tiny\% Gain Over the Baseline},
        y label style={font=\normalsize, at={(0.1,0.5)}},
        symbolic x coords={1(F), 2(F), 3(F), 4(F), 5(F), 6(F), 7(F), 8(F), 1(V), 2(V), 3(V), 4(V), 5(V), 6(V)},
        x tick label style= {
                font=\scriptsize
            },
            y tick label style= {
                font=\scriptsize
            },
        xtick = data,
        scaled y ticks = false,
        enlarge x limits=0.25,
        ymin=0,
        ymax=90,
        legend cell align=left,
        every node near coord/.append style={font=\tiny},
        legend style={
                at={(1.0,1.0)},
                anchor=north east,
                column sep=1ex
        },
    ]
    
    	\addplot[nodes near coords= P, style={rred,fill=rred,mark=none}]
             coordinates {(1(F), 28) (2(F), 23) (3(F), 65) (4(F), 48) (5(F), 53) (6(F), 27) (7(F), 16) (8(F), 50) (1(V), 51) (2(V), 80) (3(V), 52) (4(V), 28) (5(V), 44) (6(V), 48)};
        \addplot[nodes near coords=A , style={lightgray,fill=lightgray,mark=none}]
            coordinates {(1(F), 57) (2(F), 3) (3(F), 15) (4(F), 8) (5(F), 26) (6(F), 4) (7(F), 16) (8(F), 17) (1(V), 33) (2(V),16) (3(V), 8) (4(V), 8) (5(V), 44) (6(V), 28)};

        \legend{\normalsize \ouralgo, \normalsize ArduSoar}
    \end{axis}
\end{tikzpicture}
\vspace{-0.1in}
\caption{\small Relative gains in flight time over the baseline (Equation \ref{eq:base_corr}) for \ouralgo\ and ArduSoar across 14 flights at the Valley(V) and Field(F) test sites. \ouralgo\ outpeformed ArduSoar overall, yielding higher gains in 11 flights, losing in 1, and ending 2 in a draw.}
\label{fig:live_results}
\vspace{-0.22in}
\end{figure*}

\noindent
\textbf{Location.} Flights took place at two test sites denoted Valley and Field located in the Northwestern USA around 47.6\textdegree N, 122\textdegree W approximately 16km apart, each $\approx 700$m in diameter. Figure \ref{fig:test_sites} shows their layout, waypoint tracks, and geofences. \\
\backupsmall

\noindent
\textbf{Flight conditions.}
All flights for this experiment took place between October 2017 and January 2018. Regional weather during this season is generally poor for thermalling, with low daily temperatures and their amplitudes (a major factor for  thermal existence and strength), nearly constant cloud cover, and frequent gusty winds and rain. All flights were carried out in dry but at best partly cloudy weather, at temperatures $\leq 14$\textdegree C and daily temperature amplitudes $\leq 9$\textdegree C. All took place in predominant winds between 2 and 9 m/s; in $\approx 25$\% of the missions, the wind was around 7 m/s -- 78\% of our \suav{s'} 9 m/s airspeed during thermalling and off-motor glides. \\
\backupsmall

\noindent
\textbf{Flight constraints.} Due to regulations, all flights were geofenced as shown in Figure \ref{fig:test_sites} and restricted to low altitudes: 180m AGL at the Valley and 160m AGL at the Field site. To avoid collision with ground obstacles, the minimum autonomous flight altitude was 30m and 50m AGL, respectively.\\
\backupsmall

\noindent
\textbf{Mission profile.} Each mission consisted in two Radian Pros, one using \ouralgo\ and another using ArduSoar as thermalling controller, taking off within seconds of each other and repeatedly flying the same set of waypoints at a given test site counterclockwise from the Home location as long as their batteries lasted, deviating from this path only during thermal encounters. Figure \ref{fig:thermalling_alt} explains the mission pattern.

Each flight was fully autonomous until either the \suav{'s} battery voltage, as reported in the telemetry, fell below 3.3V/cell for 10s during motor-off glide, or for 10s was insufficient to continue a motor-on climb. At that point, the mission was considered over. A human operator would put the \suav{} into the Fly-By-Wire-A mode and land it.\\
\backupsmall

\noindent
\textbf{Eliminating random factors.} Live evaluation of a component as part of an end-to-end system is always affected by factors exogenous to the component. 
The \extradata{} details the measures we took to eliminate many of these factors, including the presence of non-thermal lift, potential systematic bias due to minor airframe and battery differences, and high outcome variance due to chance of finding thermals.\\
\backupsmall

\noindent
\textbf{Performance measure.} We compare \ouralgo and ArduSoar in terms of  the relative increase in flight duration they provide. Namely, for a flight at test site $S$ where controller $C^{th}$ ran on airframe $A$ and used battery $B$, we compute
\backupsmall

\begin{equation}
\textit{RelTimeGain}_{A, B, S}(C^{th}) = \frac{\textit{FlightTime}_{A,B,S}(C^{th})}{\textit{BaselineFlightTime}_{A, B, S}}
\label{eq:base_corr}
\end{equation}

\noindent
The $\textit{BaselineFlightTime}_{A, B, S}$ values are averages over durations of a series of separate flights in calm conditions \emph{with thermalling controllers turned off} (see the \extradata{}). 

\begin{center}
\textbf{Results}
\end{center}
\vspace{-0.05in}

Our comparison is based on 14 two-\suav{} thermalling flights performed in the above setup. Figure \ref{fig:live_results} shows the results. 
Importantly, they are primarily indicative of controllers' advantages over each other, rather than each controller's absolute potential to extend flight time, due to altitude and geofence restrictions that often forced \suav{s} out of thermals, and the deliberate lack of a thermal finding strategy. 

Qualitatively, as Figure \ref{fig:live_results} indicates, \ouralgo\ outpeformed ArduSoar in 11 out of 14 flights, and did as well in 2 of the remaining 3. Moreover, the flight time gains \ouralgo\ provides are appreciably larger compared to ArduSoar's. The data in the ``Effect of baseline correction'' subsection of the \extradata{} adds more nuance. It shows that without baseline correction (Equation \ref{eq:base_corr}), the results look somewhat different, even though \ouralgo\ still wins in 11 flights.

These results agree with our hypothesis that in turbulent low-altitude thermals, POMDP-driven exploration and action selection are critical for taking advantage of whatever lift there is. Likely due to active exploration, \ouralgo's trajectories can be much "messier" than ArduSoar's; see Figure \ref{fig:test_sites}.

%% file: conclusions.tex
\section{Conclusion}

This paper has presented a POMDP-based approach to autonomous thermalling. Viewing this scenario as a POMDP has allowed us to analyze it in a principled way, to identify the assumptions that make our approach feasible, and to show that this approach naturally makes deliberate environment exploration part of the thermalling strategy. This part is missing from prior work but is very important for successful thermalling in turbulent low-altitude conditions, as our experiment have demonstrated. We have presented a light-weight thermalling algorithm, \ouralgo, deployed it onboard a \suav{} equipped with a computationally constrained Pixhawk flight controller, and conducted an extensive field study of its behavior. Our experimental setup is easily replicable and minimizes the effect of external factors on soaring controller evaluation. The study has shown that in challenging low-altitude thermals, \ouralgo\ outperforms the soaring controller included in ArduPlane, a popular open-source \suav{} autopilot. Despite encouraging exploration, due to Pixhawk's computational constraints \ouralgo\ makes many approximations. However, companion computers on larger \suav{s} may be able to run a full-fledged POMDP solver in real time. Design and evaluation of a controller based on solving the thermalling POMDP near-optimally, e.g., using an approach similar to \citet{slade-iros17}'s, is a direction for future work. \\
\backupsmall

\noindent
\textbf{Acknowledgements.} We would like to thank Eric Horvitz, Chris Lovett, Shital Shah, Debadeepta Dey, Ashish Kapoor, Nicholas Lawrance, and Jen Jen Chung for thought-provoking discussions relevant to this work.

%% file: appendix.tex
\section{APPENDIX}

\input{implementation_appendix}
\input{experiments_appendix}

%% file: implementation_appendix.tex
\subsection{Implementation Details}

To make \ouralgo's implementation configurable, we have added several parameters used by \ouralgo\ to ArduPlane's SOAR\_* parameter group. They have an effect only if thermalling functionality is turned on by setting the SOAR\_ENABLE parameter to 1 \cite{tabor-arxiv18}. The new SOAR\_POMDP\_ON parameter determines whether the \suav\ should use \ouralgo\ or ArduSoar for thermalling. Similar to ArduSoar adapting ArduPilot's LOITER mode for thermalling, we implemented \ouralgo\ in Fly-By-Wire-B (FBWB) mode. This mode is similar to AUTO, but doesn't try to follow waypoints. In particular, \ouralgo\ relies on the fact that ArduPlane in FBWB will control the pitch and yaw channel to execute a coordinted turn if a roll angle (\ouralgo's output) is provided as input.

\ouralgo's performance in practice critically relies on the match between the \emph{actual} turn trajectory (a sequence of 2-D locations) the \suav{} will follow for a commanded bank angle and the turn trajectory \emph{predicted by our controller} for this bank angle and used by \ouralgo\ during planning. To predict the sequences of locations corresponding to different actions' trajectories, our implementation uses differential Equations \ref{eq:roll_damping} through \ref{eq:pos_update}:

\begin{align}
\LPdamp &= -\Kdamp\frac{ \Clp \dot{\roll}}{2\airspeed} \label{eq:roll_damping}
\end{align}
\begin{align}
\ddot{\roll} &= \frac{\Kaileron \aileronout - \LPdamp}{\Iroll} \label{eq:roll_accel}
\end{align}
\begin{align}
\dot{\yaw} &=  \frac{\gravity \tan \roll}{\airspeed} \label{eq:turn_rate}
\end{align}
\begin{align}
\dot{\uavpos} &= \left( \airspeed \sin\yaw, \airspeed \cos\yaw\right)\label{eq:pos_update}
\end{align}

\noindent
The meanings of the variables and the constants in these equations, as well as the constants' corresponding parameter names in ArduPlane and their values for Radian Pro airframes, are as follows:

\begin{itemize}
\item \yaw is the aircraft heading in radians,
\item \roll is the aircraft bank angle in radians,
\item \LPdamp is the roll damping moment,
\item $\Iroll$ is the moment of inertia about the aircraft roll axis. In ArduPlane, the corresponding parameter we have added is SOAR\_I\_MOMENT. For our Radian Pro \suav{}, $\Iroll = 0.00257482 = m_{\mathrm{radian}}k^2$,  where $m_{\mathrm{radian}}=1.2$kg is the measured mass of our modified Radian Pro and $k=0.05$ is estimated from the distribution of area in the lateral cross section.
\item \airspeed is airspeed in m/s.
\item $\Clp$ is the roll damping derivative. In ArduPlane, the corresponding new parameter is SOAR\_ROLL\_CLP. For a Radian Pro,  $\Clp = -\frac{1}{4} \frac{\pi AR}{\sqrt{\frac{AR^2}{4}+4}+2} = -1.12808704$, where $AR=11.8645073263$ is the wing aspect ratio calculated from dimensional measurements of the Radian Pro's wing.

\item $\Kdamp$ is a coefficient accounting for the wing span and other aerodynamic effects on roll damping. ArduPlane's corresponding parameter is SOAR\_K\_ROLLDAMP. For a Radian Pro, $\Kdamp=0.41073588$.

\item $\aileronout = f_{\phi_{\mathrm{PID}}}\left(\rollaction - \roll \right)$, the aileron deflection signal $\in[-1,1]$, where $f_{\phi_{\mathrm{PID}}}$ is ArduPlane's roll channel PID controller function and \rollaction is the commanded bank angle for the action.

\item $\Kaileron$ is a coefficient accounting for the rolling forces due to aileron deflection. In ArduPlane, the respective parameter is SOAR\_K\_AILERON. For a Radian Pro, $\Kaileron=1.448331$.

\item \gravity is the acceleration due to gravity.
\end{itemize}

\noindent
Note that \LPdamp and \aileronout depend on ArduPlane's PID controller parameters, and $\Iroll, \Clp, \Kdamp, \Kaileron$ are specific to the airframe type. The latter constants need to fitted to data gathered by entering the \suav{} into turns at various bank angles.

In order to predict a sequence of locations that constitute an action's trajectory given the \suav{'s} current state, our thermalling controller implementation solves Equations \ref{eq:roll_damping} to \ref{eq:pos_update} using numerical integration starting at that state, with a time step of 0.02s for the action's duration. In the explore mode, an action's duration is given by ArduPlane's SOAR\_POMDP\_HORI parameter; in the exploit mode, it is determined by a product of ArduPlane's parameters, SOAR\_POMDP\_HORI $\cdot$ SOAR\_POMDP\_EXT. In our experiments, we used SOAR\_POMDP\_HORI = 4s and SOAR\_POMDP\_EXT = 3.   
 Recall that the \suav{'s} current location -- the starting location of every trajectory -- is (0,0). When generating an action trajectory for planning, our implementation records \roll and \uavpos, the \suav{'s} (simulated) location in the air mass's reference frame, every 0.2s along the simulated trajectory. The oversampling matches the 50Hz update rate of the ArduPlane control loop, which is required for output equivalence from the simulation PID function. It also reduces error and avoids instability arising from numerical integration. The resulting location sequence for the given action is then passed to \ouralgo.

\begin{figure}
\centering
\includegraphics[trim={0 0 0 0.47cm},clip,width=0.47\textwidth]{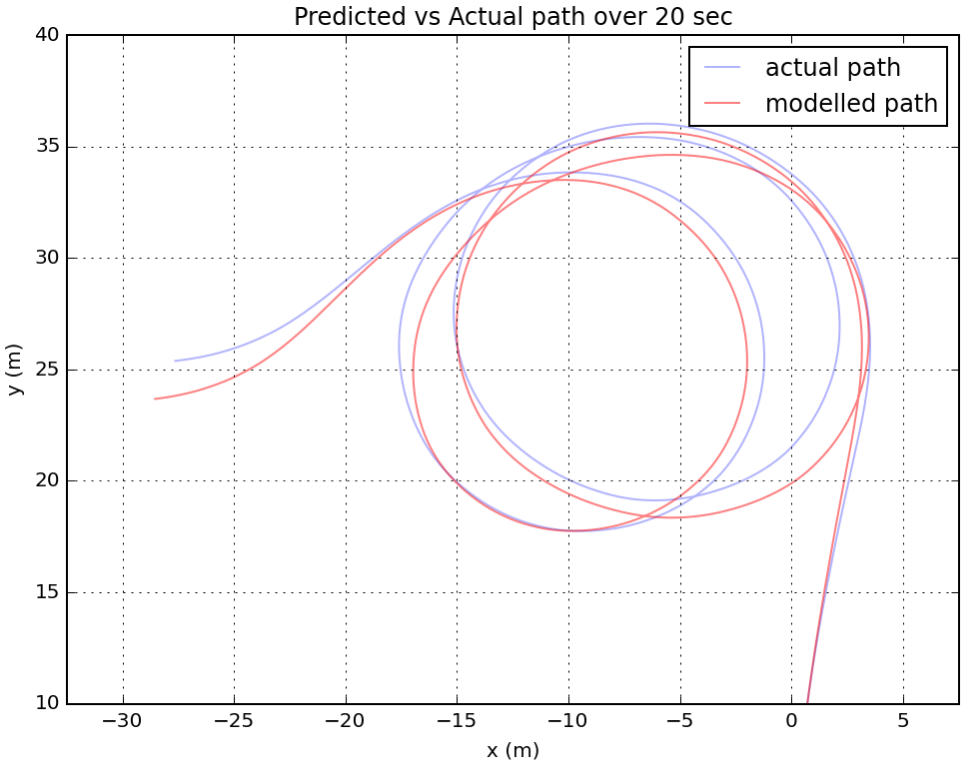}
\caption{\small Comparison between actual flight data and its predictions by the simulation model used in \ouralgo\ for the same input roll commands, over 20s of flight. 
\label{fig:predictedpath}}
\vspace{-0.25in}
\end{figure}

With the above method, we have achieved a difference of $< 1$ meter between the predicted and actual trajectories. Figure \ref{fig:model_match} shows the  match between the model-predicted and actual bank and aileron deflection angle evolutions for several roll commands, and Figure \ref{fig:predictedpath} shows the match between predicted and actual multi-action trajectories typically generated during thermalling.

Importantly, depending on the chosen range of target bank angles that encode the action set, some of these bank angles may sometimes not be achievable by ArduPlane due to its overly conservative built-in stall prevention mechanism. E.g., if the commanded bank angle is $45^{\circ}$, ArduPlane's stall prevention may override it and only allow the \suav{} to bank at $40^{\circ}$, which will cause a discrepancy between the predicted action trajectory and the actual one. In order to avoid this problem, we have introduced the SOAR\_NO\_STALLPRV parameter to ArduPlane. When set to 1, it turns off stall prevention during thermalling. Using it requires care, since stalls resulting from extreme bank angles can be very difficult to recover from. 

%% file: experiments_appendix.tex
\subsection{Experiment Methodology Details and Additional Results}

As described in the ``Empirical Evaluation'' section, our live experiment consisted in flying two identically Radian Pro airframes (one running \ouralgo, another running ArduSoar) simultaneously along the same set of waypoints. In order reduce randomness and potential bias in the outcomes of these flights, we have taken measures to eliminate the following confounding factors from our evaluation:

\begin{itemize}
\item{} \emph{Systematic bias due to airframe differences}. Despite undergoing the same tuning procedure, one airframe may perform systematically worse than the other due to minor physical differences such as discrepancies in the center of gravity positions, AHRS calibration results, current draw of the electronic equipment (especially the motor), and even dents in the airfoil.

To minimize their effect, after each flight that counted towards this experiment's results we switched the assignment of thermalling controllers to airframes. At each test site, \ouralgo\ and ArduSoar completed an equal number of experimental flights on each \suav. \\

\begin{table*}[tb]
\begin{center}
\setlength\tabcolsep{4pt}
\begin{tabular}{|c?c|c|c|c|c|c|c|c?c|c|c|c|c|c|}
\hline
Flight ID & 1(F) & 2(F) & 3(F) & 4(F) & 5(F) & 6(F) & 7(F) & 8(F) & 1(V) & 2(V) & 3(V) & 4(V) & 5(V) & 6(V) \\ \hline 
Airframe with \ouralgo & YT & BT & YT & YT & BT & YT & BT & BT & BT & YT & YT & YT & BT & BT \\ \hline 
\thead{BT flight time, minutes \\ (baseline in parentheses)} & 33(\textit{21}) & 37(\textit{30}) & 30(\textit{26}) & 27(\textit{25}) & 46(\textit{30}) & 27(\textit{26}) & 35(\textit{30}) & 39(\textit{26}) & 41(\textit{27}) & 29(\textit{25}) & 27(\textit{25}) & 27(\textit{25}) & 39(\textit{27}) & 37(\textit{25}) \\ \hline 
\thead{YT flight time, minutes \\ (baseline in parentheses)} & 32(\textit{25}) & 31(\textit{30}) & 48(\textit{29}) & 40(\textit{27}) & 38(\textit{30}) & 37(\textit{29}) & 35(\textit{30}) & 34(\textit{29}) & 36(\textit{27}) & 45(\textit{25}) & 38(\textit{25}) & 32(\textit{25}) & 39(\textit{27}) & 32(\textit{25})\\ \hline
\end{tabular}
\end{center}
\caption{\emph{Raw flight times for the Valley (V) and Field (F) test site for the YellowTail (YT) and BlackTail (BT) \suav\ airframes. Each airframe performed an equal number of flights at each test site running each thermalling controller (4 flights per controller per airframe at the Field, 3 per controller per airframe at the Valley). For each flight, the battery used was recorded. Baseline flight times are the flight times for the combination of that battery and that airframe at that test site without thermalling. These flight times were recorded in a series of separate flights on still days with thermalling turned off.}}
\label{t:live_results}
\end{table*}

\item{} \emph{Differences in battery properties.}  Similar to the previous factor, using batteries with the same nominal capacity may nonetheless result in different flight times under the same conditions. This is partly due to slight variability in the battery manufacturing process, and partly due to divergence of battery properties after several recharge cycles.

In order to eliminate battery performance from the picture, for each combination of airframe $A$, test site (waypoint sequence) $S$, and battery $B$ used in the experiments, on several calm days we measured flight duration with thermalling controllers \emph{turned off} and recorded the resulting flight times. For each $A$, $S$, and $B$, we call the average of these timings the \emph{baseline flight time}, denote it $\textit{BaselineFlightTime}_{A, B, S}$, and account for it in our performance measure for each thermalling flight (see Equation \ref{eq:base_corr}). In effect, each $\textit{BaselineFlightTime}_{A, B, S}$ is a flight duration estimate for a given $A$, $B$, and $S$ combination when the \suav{} is unaided by a thermalling controller. \\

\item{} \emph{Wave and other kinds of lift.} Although controllers aboard our \suav{s} were designed specifically for thermalling, they could chance upon rising air masses of different nature. The combination of hills, trees, and wind in the test site areas could occasionally give rise to orographic and wave lift. Data gathered in such conditions could the distort picture of \ouralgo's and ArduSoar's comparative ability to exploit thermals. 

To eliminate this risk, we avoided flying when we suspected wave lift at the test sites. However, given the virtual indistinguishability of orographic lift from low-altitude thermals, which tend to be irregular and turbulent, we included flights that may have involved orographic soaring into the results. \\

\item{} \emph{High outcome variance due to chance of finding thermals.} Making good use of even a single thermal could substantially increase a \suav{'s} flight time. Therefore, discrepancies in the chances of finding thermals --- a factor orthogonal to the quality of the thermalling controller itself --- could significantly affect flight duration.

The design of our mission profile (see the ``Mission profile'' subsection of the ``Empirical Evaluation'' section in the main paper) is motivated exactly by minimizing flight time variance due to this factor. The relatively small size of our test sites (both $< 700$m in diameter), the small range of altitudes, fixed waypoint courses, identical thermal mode triggering mechanisms, and simultaneous flights by both \suav{s} mean that both controllers regularly pass through the same locations \emph{at similar altitudes} within at most minutes of each other. In addition, we don't use any explicit thermal finding or altitude loss management strategies such McCready's speed-to-fly or memorizing locations of previous thermal encounters. All of this results in both \suav{s} often activating their thermalling controllers repeatedly at the same set of locations during a given flight, sometimes at the same time, indicating that they are likely catching the same thermals.

In addition, since both controllers are activated using identical conditions, we eliminated flights where one of the \suav{s} activated its thermalling controller at least once and the other didn't at all, because in those cases the difference in flight duration was purely due to the probabilistic nature of thermal detection. There were only a few such flights,
a testament to the robustness of the above mission profile, airframe tuning, and thermal mode triggering strategy. \\
\end{itemize}

\begin{center}
\textbf{Extended Results}
\end{center}

\noindent
\textbf{Qualitatively}, as Figure \ref{fig:live_results} indicates, \ouralgo\ outpeformed ArduSoar in 11 out of 14 flights in terms of the measure in Equation \ref{eq:base_corr}, and did as well in 2 of the remaining 3. Moreover, the flight time gains \ouralgo\ provides are appreciably larger compared to ArduSoar's.

Moreover, Table \ref{t:live_results} appears to indicate that in terms of \emph{win/loss} outcomes, baseline correction using Equation \ref{eq:base_corr} didn't make a difference for in any flight. Indeed, every time the \suav{} with the bigger absolute flight duration also happened to win according to Equation \ref{eq:base_corr}'s relative gain criterion. 

These results agree with our hypothesis that in turbulent low-altitude thermals, POMDP-driven exploration and action selection are critical for taking advantage of whatever lift there is. \\

\noindent
\textbf{Quantitatively}, however, Table \ref{t:live_results} shows that baseline correction makes a big difference when analyzing relative gains themselves. For instance, based on absolute numbers, one might think that during flight 1(F), the only one that ArduSoar won, ArduSoar won by only a narrow margin. However, as baseline-corrected results in Figure \ref{fig:live_results} demonstrate, this is not the case --- in that particular flight, ArduSoar's performance was significantly better than \ouralgo's. Therefore, baseline correction is important to apply in field experiments with many confounding factors, such as when comparing thermalling controllers.